%% file: main.tex
\documentclass[runningheads]{llncs}

\usepackage{eccv}

\usepackage{eccvabbrv}

\usepackage{graphicx}
\usepackage{booktabs}

\usepackage[accsupp]{axessibility}  

\usepackage{wrapfig}

\usepackage{hyperref}

\usepackage{orcidlink}
\newcommand\blfootnote[1]{%
  \begingroup\renewcommand\thefootnote{}\footnotetext{#1}\addtocounter{footnote}{-1}\endgroup
}

\begin{document}

\title{AdaDexGrasp: Adaptive Dexterous Grasping via 3D Visuo-Tactile Representation Fusion} 

\titlerunning{AdaDexGrasp}

\author{Xirui Liang\inst{1} \and
Jiaqi Liang\inst{1}$^{*}$ \and
Jingkai Xu\inst{1}$^{*}$ \and
Yuran Wang\inst{1}$^{*}$ \and
Ruochong Li\inst{2} \and
Yuanpei Chen\inst{1} \and
Masayoshi Tomizuka\inst{3} \and
Wei Zhan\inst{3} \and
Ruihai Wu\inst{3}$^{\dagger}$}

\authorrunning{X. Liang et al.}

\institute{Peking University, Beijing, China \and
The Hong Kong University of Science and Technology, Hong Kong, China \and
University of California, Berkeley, CA, USA}

\maketitle
\blfootnote{$^{*}$Equal contribution.\quad $^{\dagger}$Corresponding author.}

\input{sec/0_abstract}

\input{sec/1_intro}

\input{sec/2_Related_Work}

\input{sec/3_Method}
\input{sec/4_Experimentals}
\input{sec/5_Conclusions_and_Discussions}
%

\bibliographystyle{splncs04}
\bibliography{main}

\newpage
\appendix
\section*{Supplementary Material}
\addcontentsline{toc}{section}{Supplementary Material}

To further demonstrate the robustness and generalization capabilities of our method, this supplementary material provides extensive qualitative results, visualizing both grasp pose generation (Sec.~\ref{sec:supp-direct}) and the adaptive adjustment process (Sec.~\ref{sec:supp-adapt}) following initial failures across a wider range of objects in the simulation and real-world settings.
Note that the real-world results presented here utilize the robot's right arm, in contrast to the left arm used in the main text, which validates that our method is agnostic to the manipulator's specific mounting position and kinematic configuration (left vs. right), further confirming its capability to achieve stable grasping independent of the hardware setup.

\section{Visualizations of Grasp Pose Generation}
\label{sec:supp-direct}
\begin{figure*}[htbp]
  \centering
  \includegraphics[width=\textwidth]{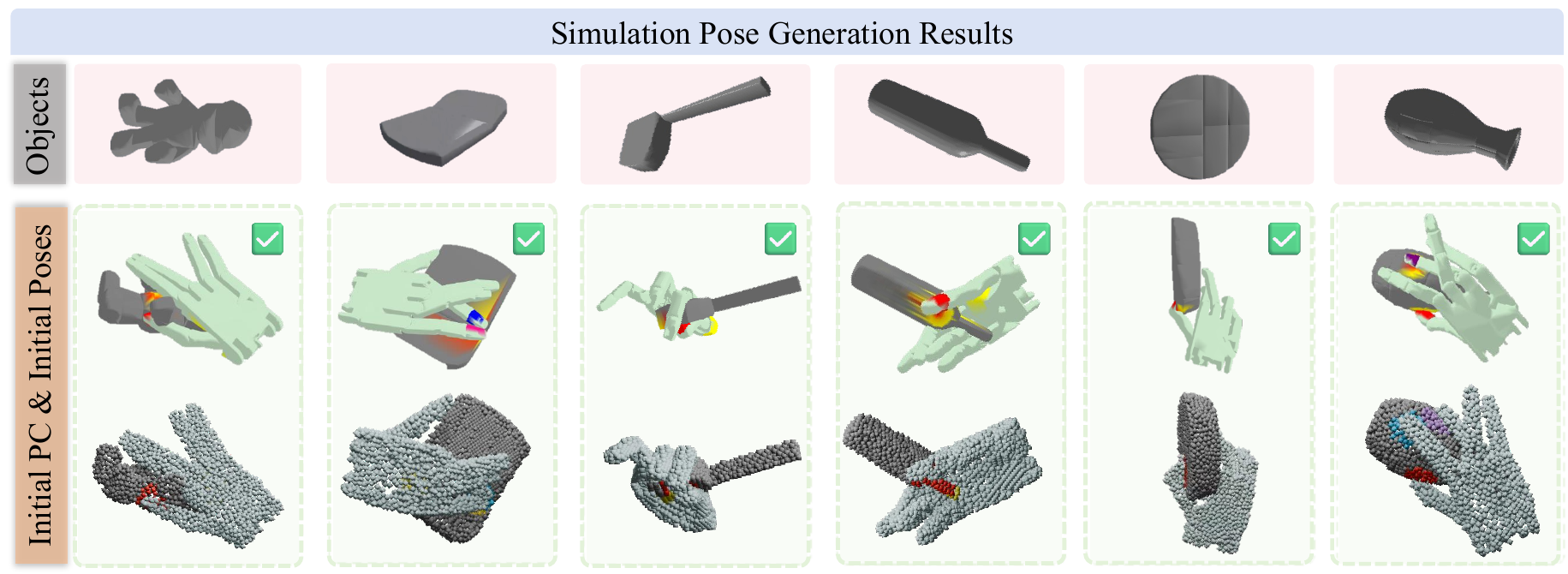}
  \caption{Visualization of successful grasp pose generation in simulation experiments.}
  \label{fig7:supp_initial_pose}
\end{figure*}

\begin{figure*}[htbp]
  \centering
  \includegraphics[width=\textwidth]{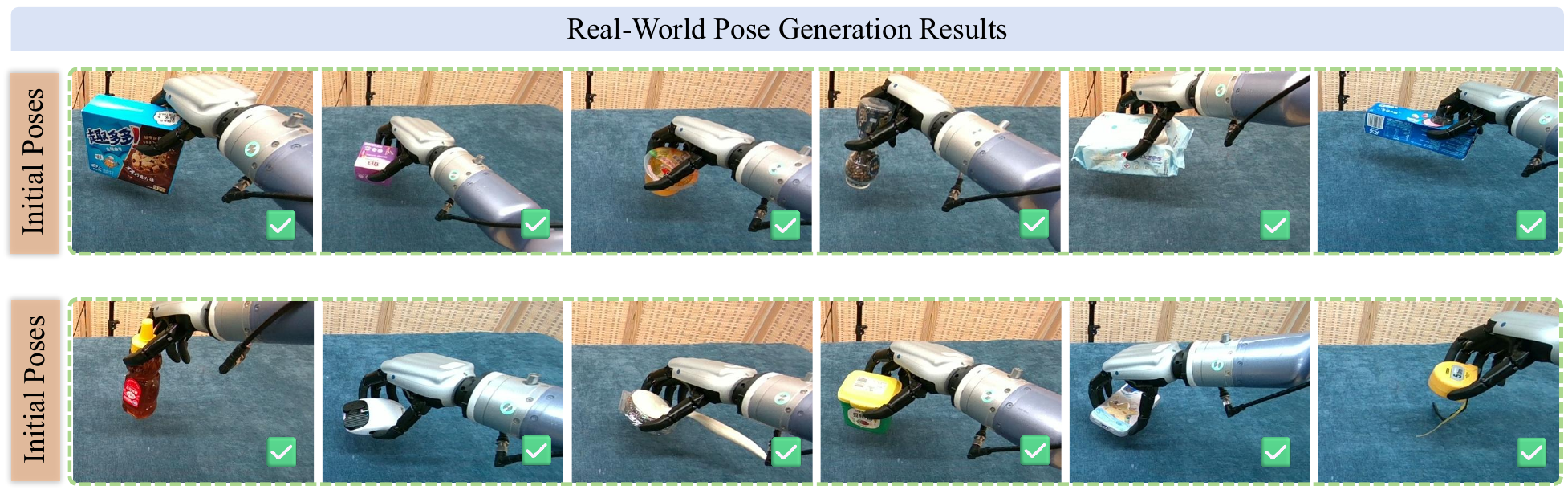}
  \caption{Visualization of successful grasp pose generation in real-world experiments.}
  \label{fig8:supp_real_initial_pose}
\end{figure*}

We present more visualizations of the generated grasping poses by our method on various challenging objects in Figure~\ref{fig7:supp_initial_pose} and Figure~\ref{fig8:supp_real_initial_pose}. The Contact-Driven Grasp Pose Generator is able to produce reasonable and stable grasping poses for complex and irregular objects such as the toy figure (1st col in Figure~\ref{fig7:supp_initial_pose}), the tape measure (3rd col in Figure~\ref{fig7:supp_initial_pose}), and the brush (2nd row, 3rd col in Figure~\ref{fig8:supp_real_initial_pose}).

\section{Training and Inference Details}
\label{sec:supp-train}
We train four models in our pipeline---the CMap generator, pose generator, classifier, and adapter---all on a single NVIDIA RTX~4090 GPU (24\,GB), each fully utilizing the available memory. The CMap and pose generators both use a batch size of 32 and a learning rate of $1\times10^{-3}$, taking about 3 hours each; the classifier uses a batch size of 128 and a learning rate of $3\times10^{-4}$ (around 10 hours); and the adapter uses a batch size of 256 and a learning rate of $1\times10^{-4}$ (around three days). These settings collectively ensure stable convergence across all components.

As for inference, the entire pipeline---including all four components---occupies approximately 16\,GB of GPU memory when running inside IsaacGym, and only about 2\,GB when executed on the real robot without the simulation environment.

In addition, regarding the hyper-parameter settings, during closed-loop refinement, we use a success threshold of $\tau_{succ}=0.5$ and a maximum iteration count of $K=5$. In practical evaluations, we observe that grasp refinement is triggered in roughly 30\% of the trials. Among those cases, over 90\% converge to a successful grasp within fewer than 3 refinement iterations, demonstrating both the efficiency and effectiveness of the adaptive optimization process.

\section{Real-Time Feasibility}
Due to the speed optimization of the Diffusion Policy backbone, our system frequency is 20--25\,Hz, feasible for real-time inference. Thus, if the object moves, our model can timely refine the grasp pose.

\section{Additional Quantitative Analysis}
\label{sec:supp-quant}
\textbf{Effect of the intermediate contact map.} Our generator first predicts a contact map $\mathcal{M}$ and then conditions the grasp pose on it. To assess the value of this intermediate representation, we train a variant that predicts the grasp pose directly from the object point cloud, without the contact map. This lowers the success rate to $82\%/78\%/77\%$ on the seen, unseen-object, and unseen-category splits, compared with $91\%/82\%/83\%$ for our full model. The consistent drop confirms that $\mathcal{M}$ provides a useful semantic prior that guides the generator toward feasible hand--object contacts.

\textbf{Grasp-proximity metric for adaptation pairs.} When constructing the failed--successful state pairs used to train the adaptation model, we match each failed state to its nearest successful counterpart under a combined $SE(3)$-and-joint distance. We ablate this design by matching with only the $SE(3)$ term or only the joint-angle term. Both variants degrade performance, to $82\%/75\%/61\%$ and $84\%/72\%/57\%$ respectively, compared with $91\%/82\%/83\%$ for the combined metric. Using either term alone yields degenerate pairs---similar root poses with entirely different finger configurations, or similar finger postures under inconsistent global placement---whereas the combined metric captures both where the hand is and how the fingers are configured, producing physically meaningful pairs and stable corrective targets.

\textbf{Tactile representation choice.} Here, the \emph{dense modality} replaces our six-dimensional tactile-intensity vector with a per-contact dense force-field vector, while the \emph{small-data} setting trains on $50\%$ of the full collected dataset under an identical framework. Under the small-data setting, our representation reaches $88\%$ accuracy versus $81\%$ for the dense modality, and in the real world it reaches $85\%$ versus $72\%$. Richer modalities demand substantially more data and widen the sim-to-real gap of tactile signals, making it harder to align simulation and reality; we therefore adopt the simpler representation for its efficiency and tighter simulation--real alignment.

\section{Real-World Evaluation Protocol}
\label{sec:supp-realworld}
We evaluate $26$ real-world objects, split into $10/9/7$ seen, unseen-object, and unseen-category groups, with $10$ randomized trials per object, yielding $260$ trials per method. The real-world trends mirror the simulation results: because the baselines are either open-loop or lack tactile feedback, they cannot detect or correct unstable contact once the initial pose is executed, whereas our tactile-driven closed-loop adaptation continuously monitors contact and recovers from such errors. Residual failures stem mainly from slippage during lifting, insufficient enclosure for objects near the hand's size limit, and unstable contact on small support areas.

\section{Visualizations of Grasp Pose Adaptation}
\label{sec:supp-adapt}
\begin{figure*}[htbp]
  \centering
  \includegraphics[width=\textwidth]{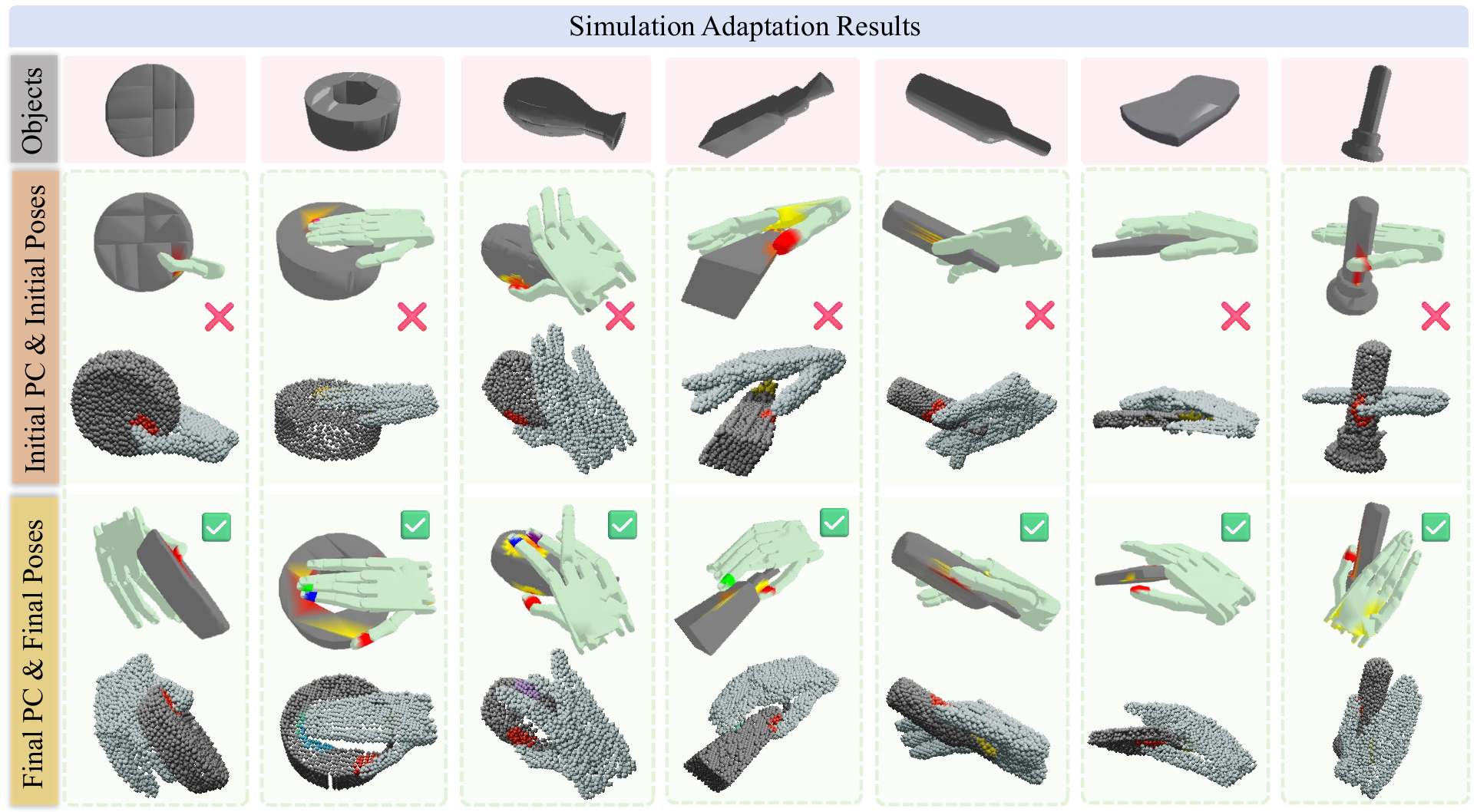}
  \caption{Visualization of grasp pose adaptation in simulation experiments.}
  \label{fig9:supp_adapt_pose}
\end{figure*}

\begin{figure*}[htbp]
  \centering
  \includegraphics[width=\textwidth]{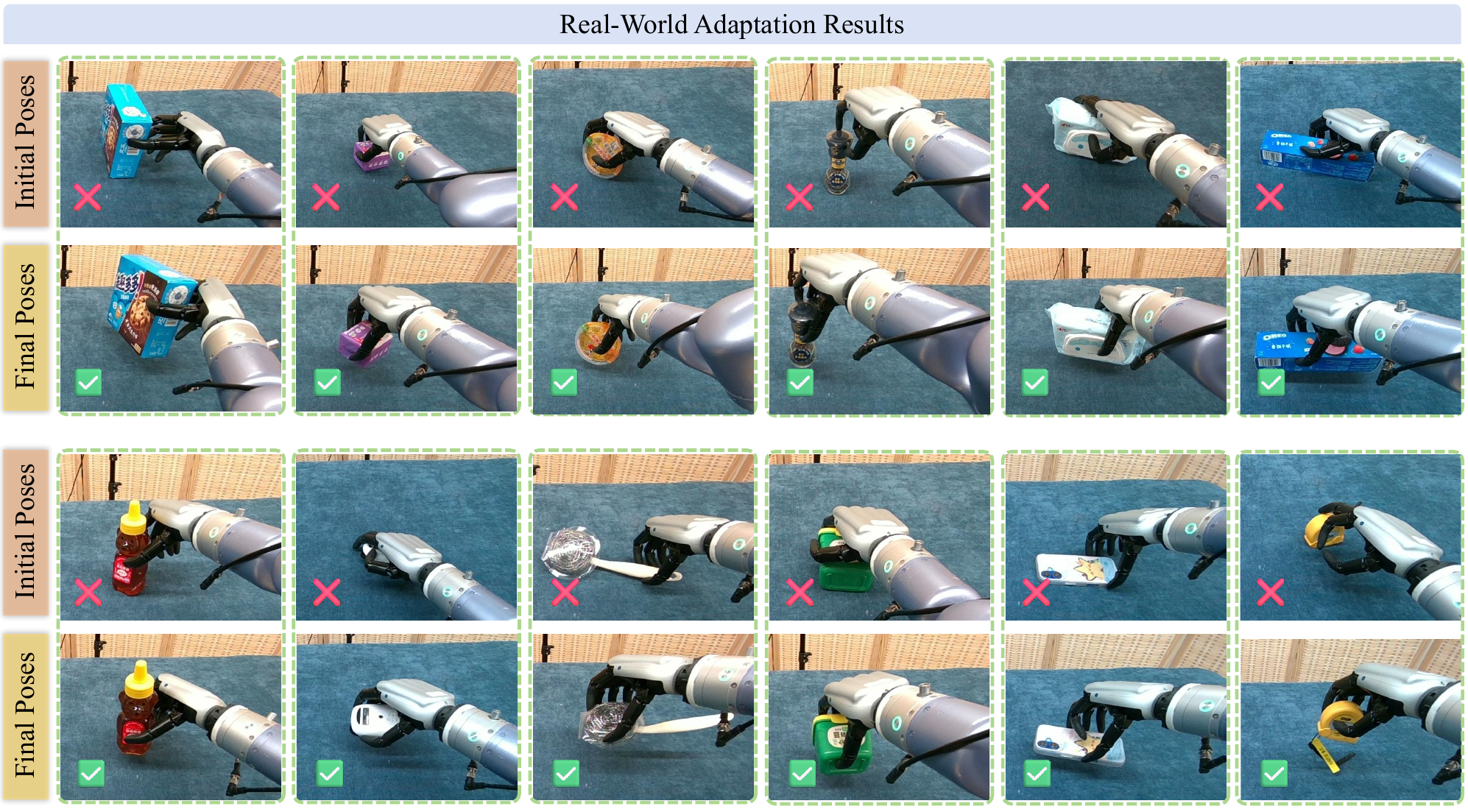}
  \caption{Visualization of grasp pose adaptation in real-world experiments.}
  \label{fig10:supp_real_adapt_pose}
\end{figure*}

We provide additional visualizations for Grasp Pose Adaptation in Figure~\ref{fig9:supp_adapt_pose} and Figure~\ref{fig10:supp_real_adapt_pose}. As observed, the initial hand poses may fail to yield a stable grasp. Common failure reasons include insufficient contact points (e.g., 3rd col in Figure~\ref{fig9:supp_adapt_pose}; 3rd row, 6th col in Figure~\ref{fig10:supp_real_adapt_pose}), limited contact surface area (e.g., 5th col in Figure~\ref{fig9:supp_adapt_pose}; 3rd row, 3rd col in Figure~\ref{fig10:supp_real_adapt_pose}), and geometric mismatch between the contact points and the object geometries (e.g., 6th col in Figure~\ref{fig9:supp_adapt_pose}; 1st row, 6th col, and 3rd row, 4th col in Figure~\ref{fig10:supp_real_adapt_pose}). However, our classifier can accurately identify these potential failure cases, while the adaptation module can correct them to a reasonable one for a successful grasp.

\end{document}

%% file: sec/0_abstract.tex
\begin{abstract}
Humans achieve stable and adaptive grasps by seamlessly integrating visual perception and tactile feedback, a capability that remains challenging to replicate in robotic systems. Existing robotic grasping approaches predominantly rely on visual inputs and lack mechanisms for tactile-guided adaptation after contact, limiting robustness and generalization. To address this challenge, we propose a unified visuo-tactile-fusion grasping framework that integrates grasp generation, feasibility prediction, and adaptive refinement. At its core, our method introduces an efficient visuo-tactile representation that tightly fuses object geometry with tactile feedback by associating tactile signals with finger identities. This unified representation supports contact-aware grasp pose generation during planning and tactile-guided refinement after contact, enabling the system to reason about fine-grained finger-object interactions and adjust grasps dynamically. Comprehensive experiments in both simulation and real-world environments demonstrate that our approach significantly enhances grasp success rates and generalization across diverse objects.

\keywords{Dexterous Grasping \and Adaptation \and Visuo-Tactile Fusion }    
\end{abstract}

%% file: sec/1_intro.tex
\section{Introduction}
\label{sec:intro}

In daily life, humans skillfully use dexterous hands to grasp a wide range of objects. This capability arises not only from the ability to plan an appropriate grasp pose based on visual perception, but also from the competence in adjusting actions in real time through tactile feedback, ensuring stable and reliable grasping. Transferring such visuo-tactile-fusion grasping capabilities to robotic systems with dexterous hands is highly important yet technically challenging.

Most existing robotic grasping methods~\cite{fang2025anydexgraspgeneraldexterousgrasping, zhang2024dexgraspnet20learninggenerative, chen2024springgraspsynthesizingcompliantdexterous, weng2024dexdiffusergeneratingdexterousgrasps, zhong2025dexgraspvlavisionlanguageactionframeworkgeneral, wan2023unidexgraspimprovingdexterousgrasping, ye2025dex1b, wei2024d} rely purely on visual information (e.g. RGB, point cloud) for planning. 
Such approaches determine the hand pose based on an initial observation, without the ability to refine or adapt the grasp according to feedback after contact. Some works~\cite{Lee_2024, guzey2023dexteritytouchselfsupervisedpretraining} explore tactile-only strategies for adaptive grasping under occlusions, but these typically require multiple trial attempts and cannot utilize available visual information, resulting in reduced efficiency.

\begin{figure*}[htb]
  \centering
  \includegraphics[width=\textwidth]{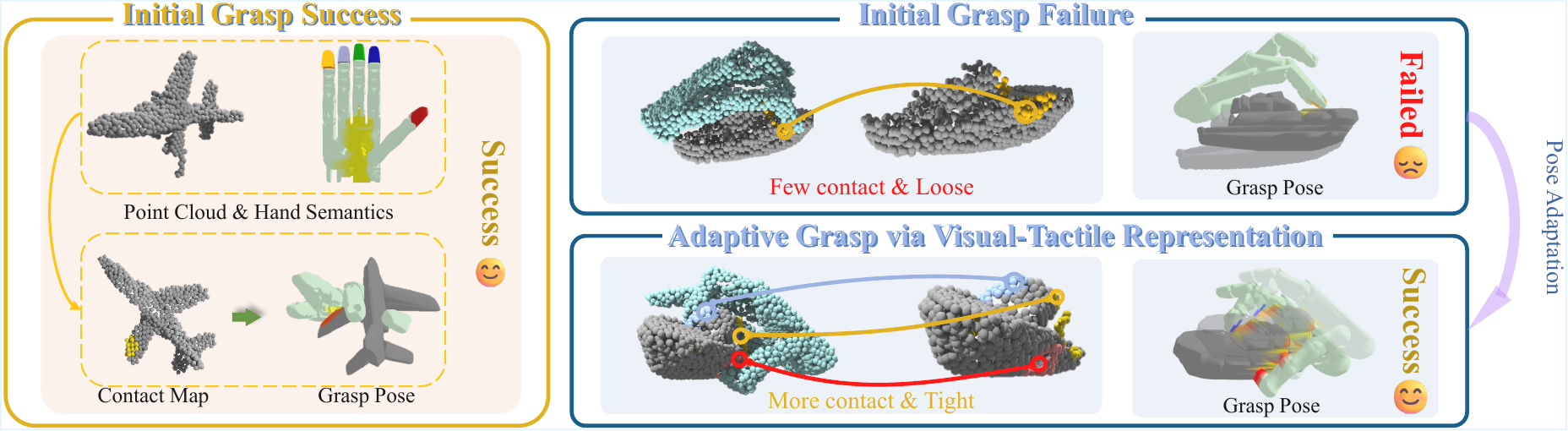}
    \caption{\textbf{Overview.}
    \textbf{(Left)} An initial grasping module predicts a contact map from the raw point cloud with finger/palm identities, aiming to generate physically successful hand poses for grasp planning informed by hand semantic information.
    \textbf{(Right-Top)} The initially generated hand pose does not guarantee a successful grasp and may fail due to sparse or unstable contact formations.
    \textbf{(Right-Bottom)} Our method further leverages a visuo-tactile-fusion representation to adapt the initial pose, resulting in more stable contact patterns that enable robust grasp execution.}
  \label{figures/Teaser}
\end{figure*}

To enhance visuo-tactile integration for improved performance, recent studies~\cite{yin2025learninginhandtranslationusing, xue2025reactivediffusionpolicyslowfast, huang20253dvitaclearningfinegrainedmanipulation, li2025adaptivevisuotactilefusionpredictive, wu2025canonicalrepresentationforcebasedpretraining} have explored a variety of multimodal fusion strategies. However, most of these efforts concentrate on object manipulation rather than achieving generalizable grasping across a wide range of categories and shapes. 
Several works~\cite{10820946, wang2025d3graspdiversedeformabledexterous, zhang2025robustdexgrasprobustdexterousgrasping} have taken initial steps toward visuo-tactile generalization in dexterous grasping. Nevertheless, their approaches mainly emphasize feature-level fusion, aiming to encourage the network to exploit both complementary and shared information from visual and tactile modalities for a richer state representation, while overlooking the spatial alignment and interaction between the two inputs, which are crucial for accurate and adaptive grasp control.

To more effectively bridge the gap between visual and tactile modalities, we propose a novel framework that integrates tactile and object geometric information to enable grasp pose generation, feasibility prediction, and adaptive refinement, which are essential for achieving efficient and robust dexterous grasping but have been largely overlooked in previous research.

Inspired by the human ability to plan which parts of the hand should interact with specific regions of an object to achieve a stable grasp, our approach introduces a more expressive contact representation during initial grasp pose generation. Unlike traditional methods that only indicate contact map without semantic information~\cite{9892548, lu2023ugg, pokhariya2024manusmarkerlessgraspcapture}, we bind contact map with finger/palm identities through tactile signals (Figure~\ref{figures/Teaser}, Top), allowing it to represent not only contact occurrence but also the expected hand–object associations. This semantic contact map is incorporated into the object point cloud features for grasp pose decoding, enabling the model to anticipate successful tactile responses during the initial planning stage and thereby improving grasp success rates.

However, for complex or unseen objects, the initially generated grasp pose may still fail to ensure a stable grasp~\cite{zhao2024graingraspdexterousgraspgeneration, Li2024ClickDiffCT}. To address this, we leverage tactile feedback obtained during execution to assess and refine the grasp (Figure~\ref{figures/Teaser}, Middle and Bottom). Our core idea is to generate tactile contact markers (which represent different contact states such as no contact or specific finger/palm contact) based on tactile signals, and integrate them into the current hand–object point cloud to create a tactile-mapped geometry representation. These point clouds are then used by a Classifier to judge the final grasp success. When a grasp is deemed suboptimal, an Adapt Model is invoked to adjust hand pose based on the current observation, aiming to achieve a successful grasp.

The absence of suitable simulation environments partly obstructs the research on visuo-tactile dexterous grasping. Consequently, we developed a tactile-equipped dexterous hand manipulation environment in IsaacGym~\cite{makoviychuk2021isaacgymhighperformance}, enabling efficient simulation-based training and evaluation. Furthermore, we also deployed our visuo-tactile-fusion method to real-world robotic systems. Comprehensive experiments were conducted to assess the generalization and robustness of our method in both simulation and real world. Both quantitative and qualitative results consistently demonstrate the effectiveness and reliability of our approach.


Our main contributions can be summarized as follows:
\begin{itemize}
    \item We introduce an effective visuo-tactile representation that tightly fuses geometric and tactile information through a contact-aware encoding, enabling fine-grained reasoning over finger–object interactions.
    \item We develop a unified visuo-tactile grasping framework that integrates grasp generation, feasibility prediction, and adaptive refinement based on tactile-guided feedback, enabling more robust and efficient dexterous grasping.
    \item Extensive experiments in both simulation and real world demonstrate the effectiveness of our framework.
\end{itemize}

%% file: sec/2_Related_Work.tex
\section{Related Work}
\label{sec:formatting}


\subsection{Dexterous Grasping}
Dexterous hand has received extensive attention for its
potential for human-like manipulation in robotics~\cite{okamura2000overview, dogar2010push, dafle2014extrinsic, nagabandi2020deep, duan2021robotics, li2016dexterous}. 
Current dexterous grasping approaches can be broadly categorized into three major paradigms: reinforcement learning (RL), imitation learning (IL), and object-centric policies.
RL methods~\cite{singh2025endtoendrlimprovesdexterous, xu2025hierarchicalreinforcementlearningarticulated,rajeswaran2017learning} acquire dexterous manipulation skills through trial-and-error exploration, while IL frameworks~\cite{alessi2025hannesimitationgraspinghannesprosthetic, wang2024genh2rlearninggeneralizablehumantorobot, wei2023wearablerobotichandhandoverhand, wu2023learning, wang2025dexgarmentlabdexterousgarmentmanipulation} transfer human demonstrations to robotic systems for efficient policy learning.
Object-centric approaches~\cite{mandikal2022dexvip, mandikal2021learning, wan2023unidexgraspimprovingdexterousgrasping}, on the other hand, leverage robot proprioception and RGB-D observations to learn visually guided grasping policies.
Despite these advancements, most existing methods rely on visual information and lack tactile perception to identify contact instabilities such as slipping or rolling on smooth surfaces~\cite{luo2017robotic}. This limitation often leads to grasp failures even when the visual geometry indicates a feasible configuration.
Recently, several studies~\cite{10820946, wang2025d3graspdiversedeformabledexterous, zhang2025robustdexgrasprobustdexterousgrasping} have begun to explore visuo-tactile integration for dexterous grasping. However, these approaches primarily focus on encoding and directly fusing the two modalities, without investigating deeper spatial and semantic relationships between vision and touch.
To address this gap, we propose a visuo-tactile grasping framework that integrates tactile feedback with object geometry, enabling more robust and efficient dexterous grasping.

\subsection{Visuo-Tactile Fusion for Manipulation}
Fusing tactile and visual information is essential for robust dexterous manipulation~\cite{lee2019makingsensevisiontouch, liu2024maniwavlearningrobotmanipulation}, as visual perception provides priors about object properties and affordances, while tactile feedback captures contact quality to guide action generation. Existing visuo-tactile fusion approaches fall into two categories: learning-based and raw-data-based methods.
Learning-based approaches typically encode each modality into latent representations and combine them through concatenation~\cite{wu2025canonicalrepresentationforcebasedpretraining} or pretraining strategies~\cite{li2019connectingtouchvisioncrossmodal, dave2024multimodalvisualtactilerepresentationlearning, 10610933, kerr2023selfsupervisedvisuotactilepretraininglocate, george2024vitalpretrainingvisuotactilepretraining, chen2022visuotactiletransformersmanipulation}. However, they mainly emphasize global feature alignment while overlooking spatial correspondence and interaction between visual and tactile inputs.
Raw-data-based methods~\cite{yin2023rotatingseeinginhanddexterity, yin2025learninginhandtranslationusing} attempt direct signal-level fusion, but they likewise fail to capture the geometric and contact-level relationships crucial for coordinated perception.
In contrast, our approach explicitly aligns modalities by integrating tactile signals into the object point cloud for grasp generation and constructing tactile-mapped geometric representations for adaptive refinement, effectively bridging visual and tactile spaces to achieve more stable, efficient, and robust dexterous grasping.

%% file: sec/3_Method.tex
\section{Method}

\begin{figure*}[htb]
  \centering
  \includegraphics[width=\textwidth]{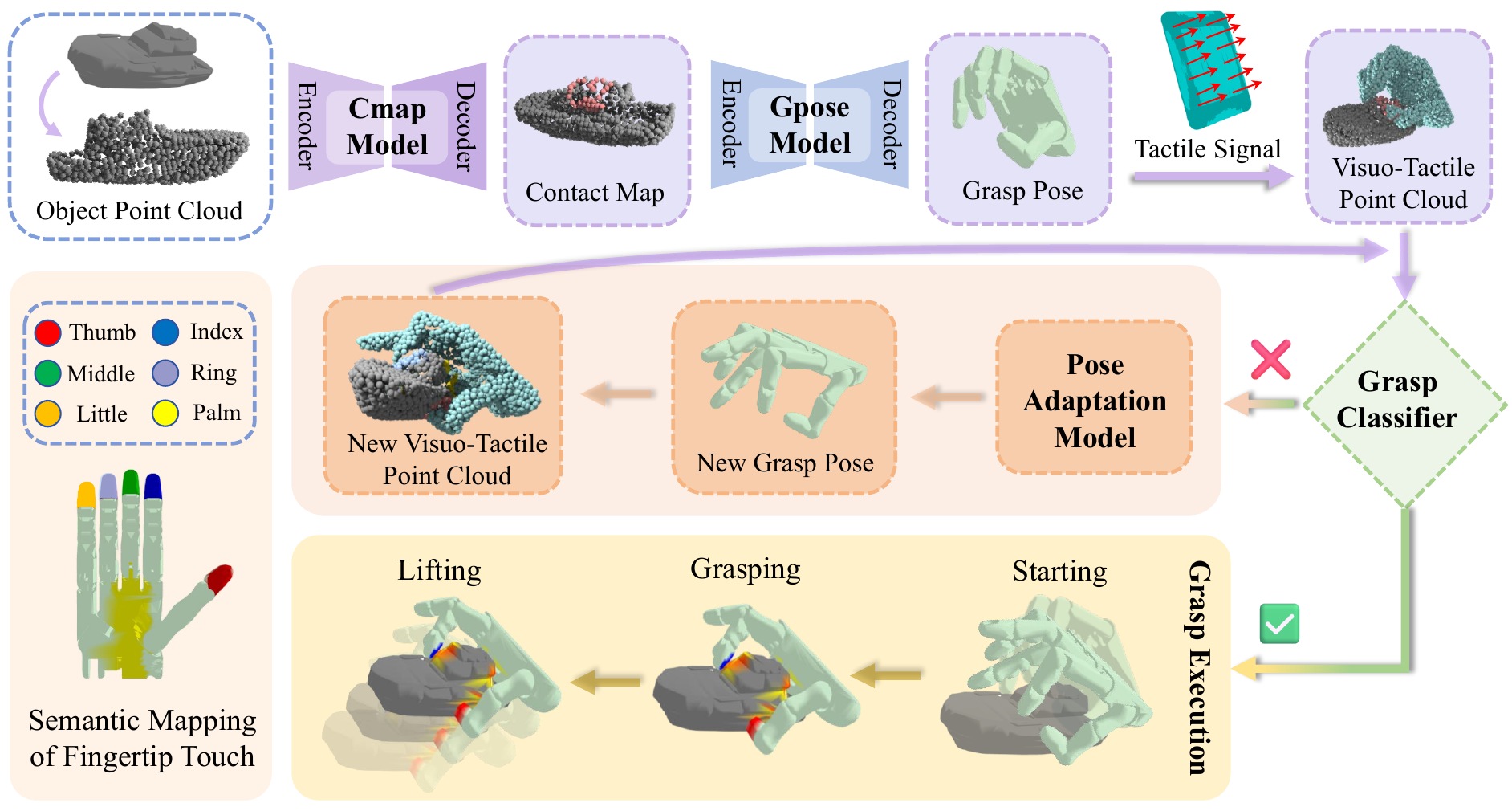}
\caption{\textbf{Pipeline of the proposed visuo-tactile grasping framework.}
Given an object point cloud and semantic fingertip labels, the system first predicts a contact map using the Cmap model and generates an initial grasp pose with the Gpose model. 
The predicted hand configuration and tactile feedback are fused into a visuo–tactile point cloud that captures both geometric and contact cues. 
A grasp classifier evaluates grasp stability, and if failure is predicted, a pose adaptation module iteratively refines the grasp by generating updated poses and visuo–tactile observations until success is predicted or a maximum number of iterations is reached. 
The final grasp is executed through a staged process of reaching, grasping, and lifting, enabling stable interaction with diverse objects.}  
  \label{fig2:pipeline}
\end{figure*}

\subsection{Problem Formulation}

We consider the problem of learning stable and adaptive grasping with a tactile-equipped dexterous hand. 
$\mathcal{O}$ denotes the target object, represented by a point cloud 
$\mathcal{P}_{obj} = \{(x_i, y_i, z_i, r_i, g_i, b_i)\}_{i=1}^{N_0}$ 
captured by an RGB-D camera sensor. 

The robot hand state is defined as $s = [p, q, \mathbf{j}]$, where 
$p \in \mathbb{R}^3$ and $q \in \mathbb{R}^4$ denote the palm position and orientation, 
and $\mathbf{j} \in \mathbb{R}^{22}$ represents the joint angles of the $22$ actuated joints. 
During interaction, tactile sensors on the fingertips and palm provide readings 
$\tau = \{\tau_i\}_{i=1}^{N_t}$ corresponding to local contact pressures, where ${N_t}$ denotes the number of tactile sensors (six in our setup: five fingertips and one palm).

Our objective is to find a grasp configuration $s^*$ that maximizes the probability of grasp success conditioned on visual and tactile observations:
\begin{equation}
s^* = \arg\max_s \; p(y=1 \mid \mathcal{P}_{obj}, \tau, s),
\end{equation}
where $y \in \{0,1\}$ indicates grasp success. 
The key challenge is modeling the relationship between visuo-tactile observations and grasp stability.

\subsection{Overview}

\begin{figure*}[htb]
    \centering
    \includegraphics[width=\textwidth]{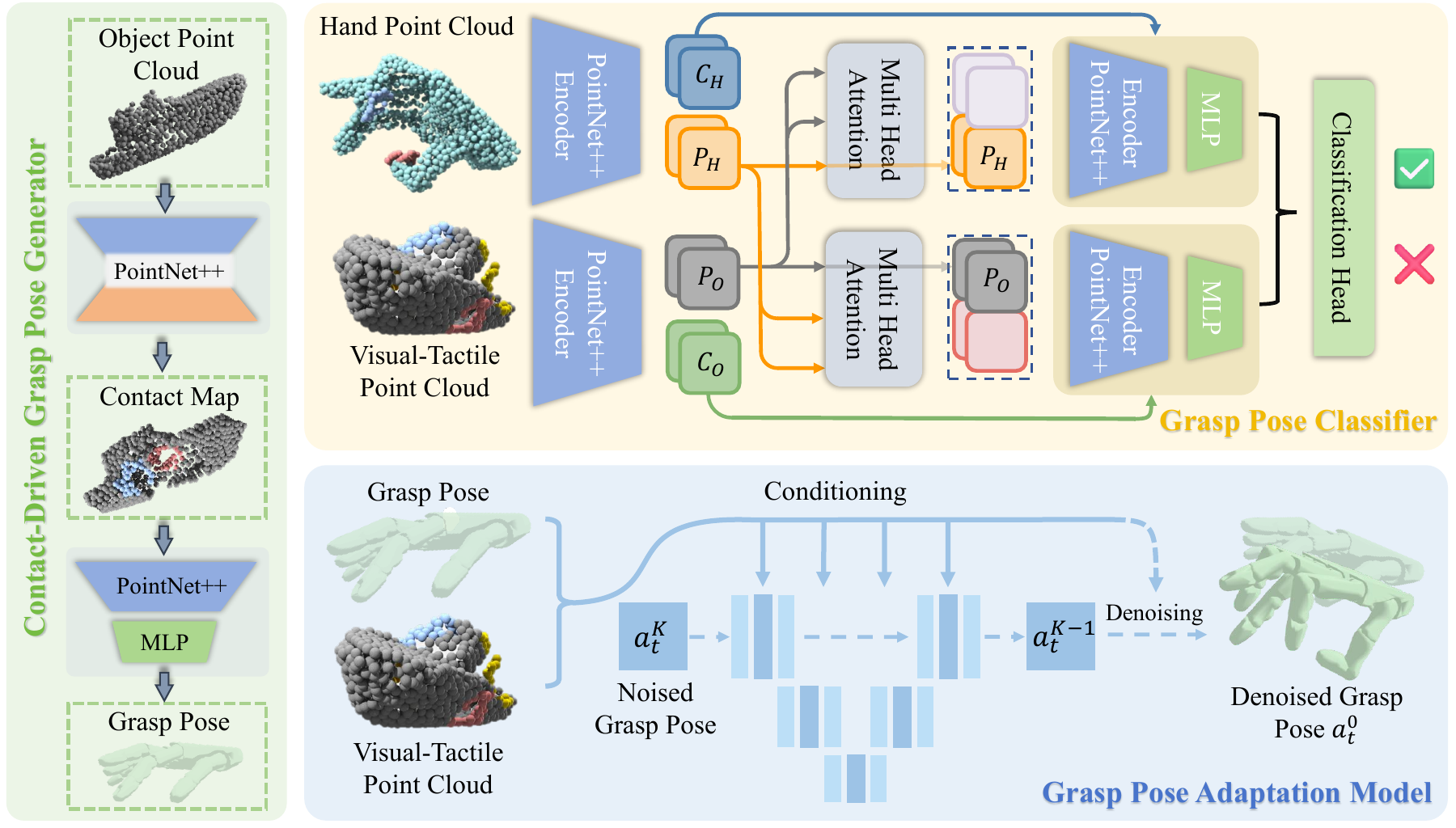}
    \caption{\textbf{Architecture of the grasping framework.} 
    The system consists of a contact-driven grasp generator, a grasp classifier, and a grasp adaptation module for stable dexterous grasping.
    \textbf{Green:} the object point cloud is processed by PointNet++ to predict a visuo–tactile contact map and generate an initial grasp pose.
    \textbf{Yellow:} a dual-encoder PointNet++ architecture encodes the hand and visuo–tactile point clouds, followed by multi-head attention for grasp success prediction. 
    \textbf{Blue:} if failure is predicted, a diffusion-based adaptation module iteratively refines the grasp pose conditioned on visuo–tactile features.}
    \label{fig3:modelArchitecture}
\end{figure*}

To address this challenge, we propose a unified framework that integrates object geometry and tactile feedback to support grasp pose generation, feasibility prediction, and adaptive refinement. The overall pipeline is illustrated in Figure~\ref{fig2:pipeline}.
The architecture of the proposed method (Figure~\ref{fig3:modelArchitecture}) consists of three components:
\begin{itemize}
    \item \textbf{Contact-Driven Grasp Pose Generator} (Section~\ref{grasp_generation}), which produces the initial grasp pose.
    \item \textbf{Grasp Pose Classifier Model} (Section~\ref{grasp_classifier}), which predicts grasp feasibility and stability.
    \item \textbf{Grasp Pose Adaptation Model} (Section~\ref{grasp_correction}), which refines suboptimal grasp poses.
\end{itemize}

\textbf{Closed-Loop Grasp Optimization} (Section~\ref{close-loop}) integrates these components into an iterative refinement process.  Section~\ref{data_collection} further describes the training data collection procedure.


\subsection{Contact-Driven Grasp Pose Generator}
\label{grasp_generation}

Given the point cloud of an object, we generate an initial grasp hypothesis by predicting a contact distribution over the object surface. 
Model $h_\psi$ predicts a probabilistic contact map $\mathcal{M}$:

\begin{equation}
    \begin{split}
        \mathcal{M} 
        &= \{ \mathcal{M}_i \mid 
        \mathcal{M}_i \in \{1,2,3,4,5,6\},\\
        &\quad i = 1, 2, \dots,N_0 \}
        = h_{\psi_1}(\mathcal{P}_{obj})
    \end{split}
\end{equation}

where each label $\mathcal{M}_i$ corresponds to a specific hand part (finger or palm). Thus, the predicted contact map encodes potential tactile interactions between the hand and the object. 

The grasp generator $h'_\psi$ then conditions on $(\mathcal{P}_{obj}, \mathcal{M})$ to predict an initial grasp state:

\begin{equation}
s^{(0)} = h'_{\psi_2}(\mathcal{P}_{obj}, \mathcal{M}).
\end{equation}

This estimate $s_0$ serves as an initial grasp pose that often succeeds directly or provides a good starting point for subsequent refinement.

\begin{figure*}[htb]
    \centering
    \includegraphics[width=1.0\textwidth]{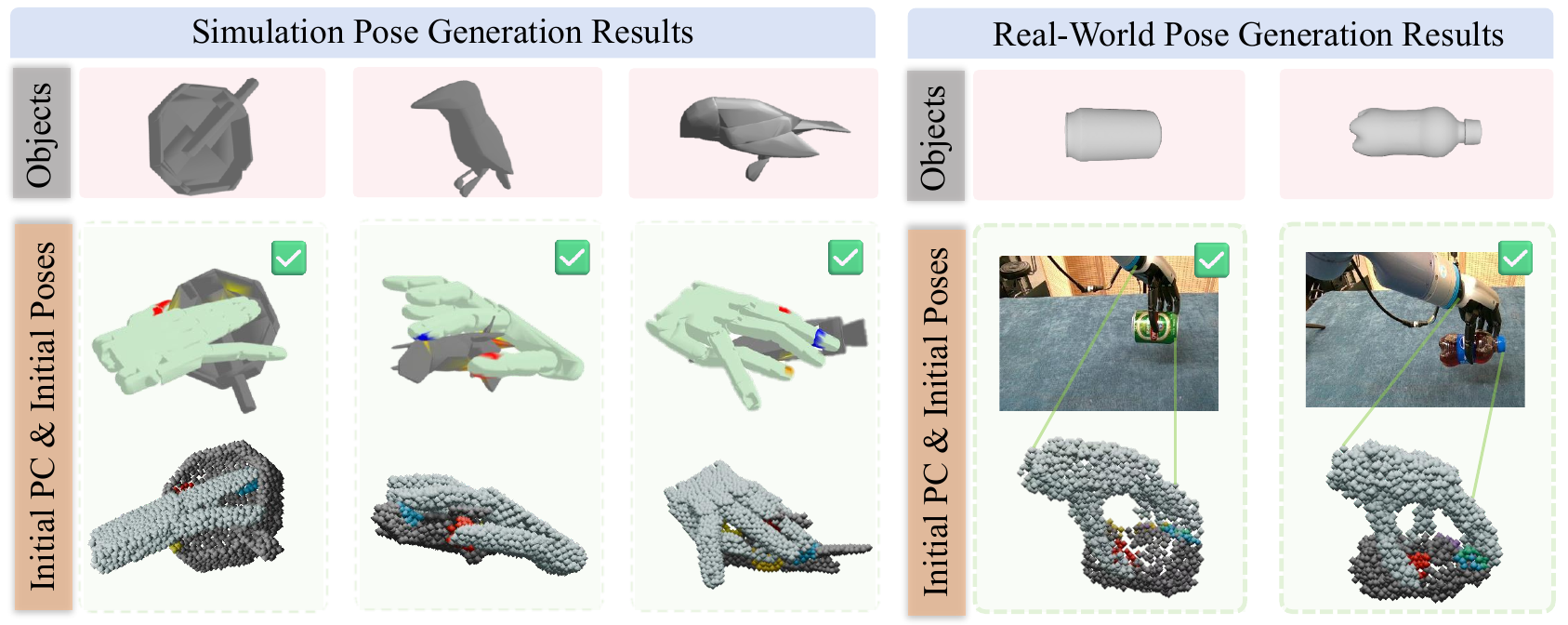}
    \caption{\textbf{Visualization of Grasp Pose Generation}. Row 1 presents object samples in both simulation and the real world. As shown in Row 2 and Row 3, the predicted contact map with semantic information (finger/palm identities) enables grasp pose generator to generate successful pose for target object grasping.}
    \label{fig:generation}
\end{figure*}

During initial grasp generation, tactile inputs are unavailable because no contact has occurred. 
Each training sample includes hand joint states, grasp success labels, and visuo–tactile point clouds. 
The contact map prediction is supervised using cross-entropy loss: 

\begin{equation}
\mathcal{L}_{CE}(\mathcal{M}, \mathcal{M}_s)= - \sum_{i} M_{s,i} \log M_i
\end{equation} 

while the grasp pose is supervised using mean-squared error:

\begin{equation}
\mathcal{L}_{MSE} = \|s^{(0)} - s^s\|_2^2
\end{equation}

Training data are collected through RL rollouts and can be adapted to different dexterous hands by training the policy on the corresponding hand models. 
Figure~\ref{fig:generation} visualizes the initial grasp generation process.

\subsection{Grasp Pose Classifier Model}
\label{grasp_classifier}
Due to the diversity of hand poses and object-dependent grasp requirements, an initially generated grasp may not always succeed. We therefore predict grasp feasibility based on the current hand–object interaction state.

After executing the initial grasp pose $s_0$, we annotate the object point cloud with tactile signals derived from contact forces at the five fingertips and the palm. Each contact region is assigned a unique ID corresponding to the contacting hand part, producing a tactile-mapped point cloud $\mathcal{P}_{vt}$.

It is worth clarifying the distinction between the contact map $\mathcal{M}$ and the tactile-mapped point cloud $\mathcal{P}_{vt}$, as both encode contact-related information but play different roles. The contact map $\mathcal{M}$ is a \emph{predicted} per-point categorical prior inferred from the object point cloud and used \emph{before} execution to guide initial grasp generation. In contrast, $\mathcal{P}_{vt}$ is the \emph{post-contact} object point cloud annotated with \emph{measured} tactile signals, consumed by the classifier and the adaptation model. 

Given $\mathcal{P}_{vt}$, the classifier estimates the probability that the current configuration yields a stable grasp. Formally, the model learns a discriminative mapping:
\begin{equation}
\hat{y} = f_\theta(\mathcal{P}_{vt}, s),
\end{equation}
where $\hat{y} \in [0,1]$ denotes the predicted grasp success probability and $f_\theta$ is a PointNet-based network. The model is trained using binary cross-entropy:
\begin{equation}
\mathcal{L}_{cls} = -[y \log(\hat{y}) + (1-y) \log(1-\hat{y})].
\end{equation}

By leveraging both geometric structure and tactile contact patterns, $f_\theta$ provides a differentiable estimate of grasp stability that can be evaluated in real time.

\subsection{Grasp Pose Adaptation Model}
\label{grasp_correction}

To recover from failed grasps, we introduce a grasp pose adaptation model $g_\phi$ that transforms unstable grasps into stable ones.

\begin{figure*}[htb]
    \centering
    \includegraphics[width=1.0\textwidth]{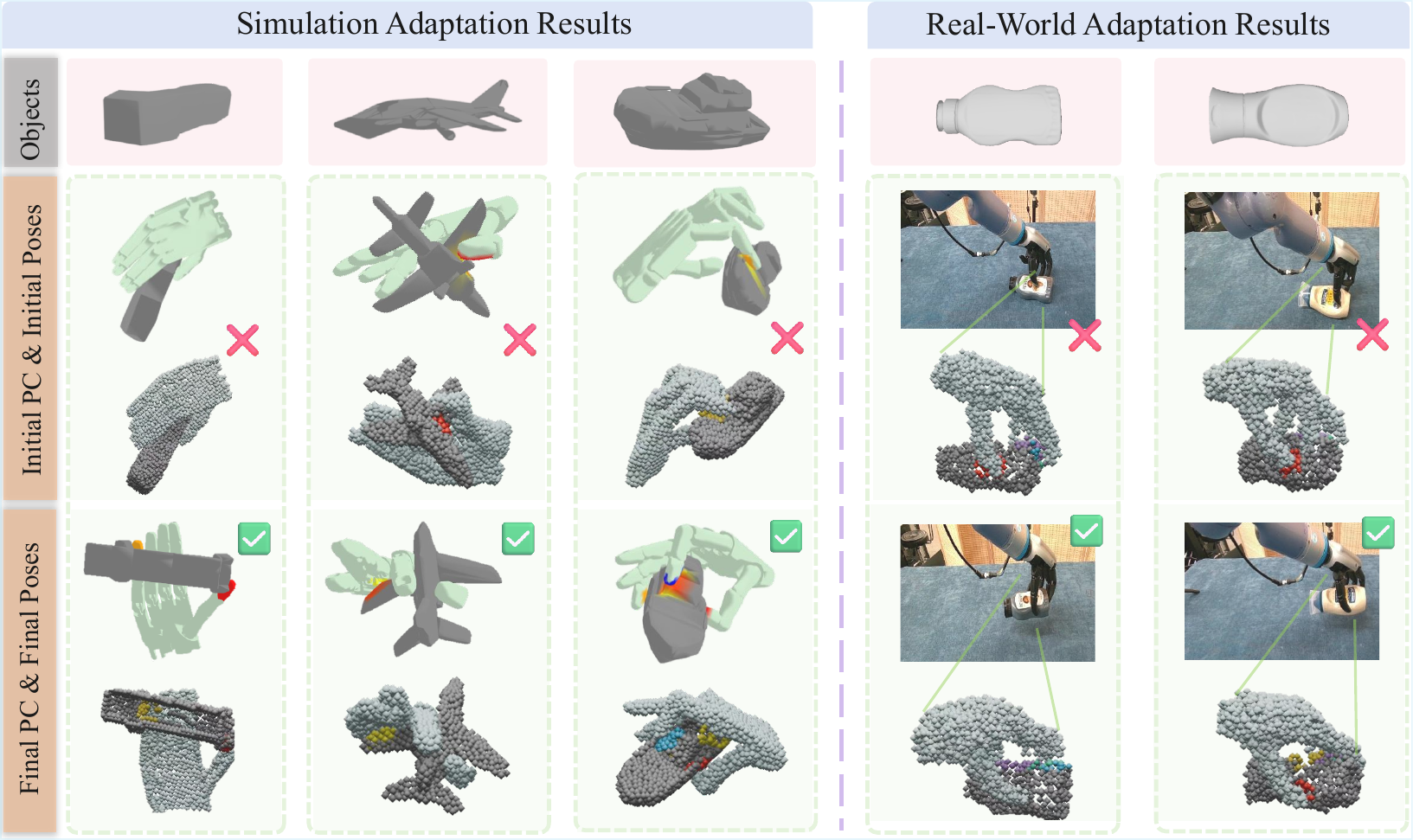}
    \caption{\textbf{Visualization of Dexterous Grasp Pose Adaptation.} Row 1 presents object samples in both simulation and the real world. As shown in Row 2 (hand–object interaction mesh) and Row 3 (hand–object interaction point cloud), the initially generated grasp poses may fail to securely grasp the object, as evidenced by sparse or weak contact regions that lead to loose or unstable grasps. Row 4 and Row 5 illustrate the refined grasp poses and updated contact states after adaptation, where hand–object interaction becomes significantly more stable, with denser contact regions and a larger support area in the point cloud. These observations further validate the effectiveness of the proposed adaptation module in correcting suboptimal initial grasps.}
    \label{fig:adaptation}
\end{figure*}

For training, we construct paired tactile-enhanced states:
\[
(\mathcal{P}_{vt}^f, s^f, y=0), \quad (\mathcal{P}_{vt}^s, s^s, y=1),
\]
where the successful grasp $s^s$ is geometrically and temporally close to the failed grasp $s^f$. Concretely, for each object we pool successful and failed states from the PPO rollouts immediately before lifting, so that all candidates share the same interaction stage, and transform them into an object-centric frame to remove any dependence on absolute object placement. Each failed state $s^f$ is then paired with its nearest successful counterpart $s^s$ under the distance
\begin{equation}
d(s, s') = 10\sqrt{\|p - p'\|_2^2 + \theta_{q,q'}^2} + \|\mathbf{j} - \mathbf{j}'\|_2,
\end{equation}
where $(p, q, \mathbf{j})$ denote the palm translation, palm orientation, and the $22$ joint angles, and $\theta_{q,q'}$ is the geodesic distance between the two orientations. 

The model learns a corrective mapping:
\begin{equation}
\Delta s = g_\phi(\mathcal{P}_{vt}^{f}, s^f),
\end{equation}

where $\Delta s$ represents adjustments to the hand pose and joint angles. The refined grasp is:
\begin{equation}
s^{f,\mathrm{new}} = s^{f} + \Delta s.
\end{equation}

The training objective encourages the corrected pose to approach the successful grasp while maximizing predicted grasp stability:
\begin{equation}
\mathcal{L}_{gen} =
\|\Delta s - (s^s - s^f)\|_2^2
+ \lambda (1 - f_\theta(\mathcal{P}_{vt}^f, s^f + \Delta s)).
\end{equation}

This objective enables the model to learn corrective actions guided by both tactile perception and grasp success prediction. Figure~\ref{fig:adaptation} visualizes the adaptation process.

\subsection{Closed-Loop Grasp Optimization}
\label{close-loop}

The contact-driven grasp generator, grasp classifier, and grasp adaptation model are integrated into a closed-loop refinement process. Starting from the initial pose $s_0$, the system iteratively evaluates and updates the grasp:

\begin{align}
\hat{y}_k &= f_\theta(\mathcal{P}_{vt}^{k}, s^{(k)}), \\
s_{k+1} &=
\begin{cases}
s_k, & \text{if } \hat{y}_k > \tau_{succ}, \\
s_k + g_\phi(\mathcal{P}_k, s_k), & \text{otherwise},
\end{cases} \\
\text{stop if } &~ \hat{y}_k > \tau_{succ} \text{ or } k = K.
\end{align}

Here $\tau_{succ}$ is the success threshold. The process terminates when the predicted success probability exceeds $\tau_{succ}$ or the maximum iteration number $K$ is reached. The resulting grasp configuration $s^*$ satisfies both geometric feasibility and tactile stability, enabling robust grasping under diverse contact conditions.

\subsection{Tactile-Enhanced Data Collection via Reinforcement Learning}
\label{data_collection}
We collect training data using reinforcement learning following the first-stage PPO setup in UniDexGrasp~\cite{xu2023unidexgraspuniversalroboticdexterous}. 
This strategy generates diverse grasp poses on the same object while maintaining stable grasps with minimal refinement.

A dexterous hand is trained to grasp across various objects using Proximal Policy Optimization (PPO)~\cite{schulman2017proximalpolicyoptimizationalgorithms}. 
Each interaction produces a trajectory
\[
\mathcal{T} = \{(s_t, a_t, r_t, \tau_t)\}_{t=1}^T,
\]
where $a_t$ and $r_t$ denote the control command and grasp reward respectively. 
Both successful and failed attempts are collected to ensure diverse tactile data.

During data collection, tactile information is annotated for each interaction by monitoring contact forces on the fingertips and palm. 
If the force on a finger exceeds a predefined threshold (0.01N in our experiments), that finger is marked as active. 
For each detected contact, we select nearby object points and assign them corresponding contact labels.

Based on these annotations, we construct a tactile-mapped point cloud $\mathcal{P}$ with contact identifiers for each hand part. 
The representation is defined as
\begin{equation}
\mathcal{P}_{vt} =
\left\{ (x_i, y_i, z_i, r_i, g_i, b_i, k_i, \mathbf{c}_i) \mid i=1,\dots,N \right\},
\end{equation}
where $(x_i, y_i, z_i)$ denote 3D coordinates and $(r_i, g_i, b_i)$ are color channels. 
The scalar $k_i \in \mathbb{R}$ indicates the contact state between the hand and the object, while $\mathbf{c}_i \in [0,1]^6$ represents a six-dimensional tactile intensity vector corresponding to the five fingertips and the palm.

This fused representation captures both geometric contact distribution and tactile intensity, and serves as the unified input for subsequent models. 
Each dataset sample is represented as $(\mathcal{P}_{vt}, s, y)$, where the binary label $y$ indicates grasp success or failure.

%% file: sec/4_Experimentals.tex
\section{Experiments}

We evaluate the proposed visuo-tactile fusion grasping framework in both simulation and real-world settings using a tactile-enabled dexterous hand. The simulation setup is described in Section~\ref{setup}, followed by baseline comparisons (Section~\ref{baseline_tex}) and ablation study (Section~\ref{ablation_tex}). Real-world experiments (Section~\ref{real-world experiment}) further validate the feasibility and practicality of the proposed approach.

\vspace{-2mm}
\subsection{Simulation Experimental Setup}
\label{setup}

Our simulation environment is built upon IsaacGym~\cite{makoviychuk2021isaacgymhighperformance}, enabling large-scale parallel collection of grasp trajectories and tactile feedback under realistic physics. The grasping task consists of three subtasks: reaching, grasping, and lifting.

\textbf{Assets.} The dexterous hand is based on ShadowHand, featuring five fingers with 22 degrees of freedom (DoF). Including the 6-DoF root pose controlling spatial motion, the system has 28 DoF in total. In simulation, the hand is moved to target positions by directly modifying the root pose.
Object assets are taken from UniDexGrasp++~\cite{wan2023unidexgraspimprovingdexterousgrasping}, including 50 objects from 6 categories. Among them, 20 objects are used for training and 30 unseen-category objects for testing. 

\textbf{Metrics.} A grasp is considered successful if the object remains stable without slippage for at least $1$ second after lift-off. We report the \textbf{grasp success rate} and evaluate performance on three object sets:
\begin{itemize}
    \item Seen Objects. Objects used in training.
    \item Unseen Objects. Objects from the same categories but unseen in training.
    \item Unseen Categories. Objects from categories absent in training.
\end{itemize}


\subsection{Baselines}
\label{baseline_tex}

We compare our method against ten state-of-the-art dexterous grasping frameworks, all baselines are evaluated under identical simulation settings using the same grasp success metric.

\begin{itemize}
    \item\textbf{DexGraspAnything}~\cite{zhong2025dexgraspanythinguniversalrobotic}: a large-scale grasp generator trained on massive grasp–object pairs, primarily exploiting visual geometry.
    \item\textbf{UniDexGrasp++}~\cite{wan2023unidexgraspimprovingdexterousgrasping}: a diffusion-based framework that predicts grasp poses for multiple robotic hands without tactile feedback.
    \item\textbf{DexDiffuser}~\cite{weng2024dexdiffusergeneratingdexterousgrasps}: a model that samples diverse grasp configurations from object point clouds.
    \item\textbf{UniDexGrasp(PPO)}~\cite{xu2023unidexgraspuniversalroboticdexterous}: a PPO policy based on UniDexGrasp using our visuo-tactile inputs.
    \item \textbf{Robot Synesthesia}: a tactile distillation approach using tactile point clouds. 
    \item \textbf{Intuitive Closed-Loop}: a heuristic closed-loop policy following the rule “move closer when no contact is detected”.
    \item\textbf{ContactDexNet}~\cite{zhang2025contactdexnetmultifingeredrobotichand}: a method that proposes grasp poses from predicted contact maps.
    \item\textbf{DexGraspVLA}~\cite{zhong2025dexgraspvlavisionlanguageactionframeworkgeneral}: a VLA-based grasping model using RGB input.
    \item\textbf{UniDexGrasp-CL}~\cite{xu2023unidexgraspuniversalroboticdexterous}: UniDexGrasp augmented with a failure classifier that resamples a new grasp whenever failure is predicted, giving it closed-loop refinement capability.
    \item\textbf{DexDiffuser-VT}~\cite{weng2024dexdiffusergeneratingdexterousgrasps}: DexDiffuser extended with tactile conditioning, where tactile features are concatenated with its visual input during evaluation and refinement.
\end{itemize}


\input{table/baseline}

From Table~\ref{baseline}, we observe that our method consistently outperforms all baseline approaches across the three object sets, demonstrating superior robustness and generalization capability.
Through analyzing the grasp poses generated by the baselines during experiments, we identify four major issues.


1) During the initial grasp pose generation, generative methods (Row 1, Row 2, Row 3) rely solely on visual geometry; without intermediate semantic correspondence, they often produce infeasible grasp configurations for complex objects, resulting in grasp failures.
2) While end-to-end models (Row 3) directly map RGB inputs to continuous actions, they lack precise understanding of geometric structures, making them fail easily when handling hard-to-grasp objects.
3) Lacking real-time tactile feedback, even offline contact modeling (Row 7) yields only static predictions. Consequently, it cannot perceive actual contact states to perform secondary closed-loop adjustments against initial spatial misalignments.
4) In the refinement stage, existing visuo-tactile methods struggle with stability. Pure RL policy (Row 4) suffers from high-dimensional exploration noise; distilled model(Row 5) remains brittle to local geometry shifts; and the Intuitive Closed-Loop policy blindly tightens its fingers toward the object without geometric awareness, inevitably leading to failure. Resampling-based refinement (Row 9, UniDexGrasp-CL) stays confined to the same generative distribution and, lacking contact information from the failed attempt, cannot escape the original failure modes; and DexDiffuser-VT (Row 10) fuses modalities weakly, as its Basis Point Set encoding captures only global object features while destroying the local point-cloud structure that tactile signals depend on.


In contrast, our approach incorporates a contact-aware representation that encodes finger and palm identities into grasp generation, along with tactile-guided iterative refinement. This enables feasible contact prediction and real-time pose correction. The high success rates on both unseen objects and unseen categories highlight the strong generalization capability of our framework, suggesting that combining visual geometry with semantic and tactile cues is critical for reliable dexterous grasping.

Furthermore, the performance gap between our method and the baselines increases for unseen categories, indicating that baselines struggle to generalize to novel object shapes and configurations. This emphasizes the importance of incorporating hand–object semantic awareness and closed-loop tactile feedback in developing robust grasping policies.
The gains also persist at scale: on the DexGrasp Anything dataset~\cite{zhong2025dexgraspanythinguniversalrobotic} (15k+ objects), our method still reaches $87\%$.

\subsection{Ablations}
\label{ablation_tex}
To investigate the contribution of each component, we conduct ablation experiments focusing on the tactile encoding and adaptation. Our ablations include:

\begin{itemize}
    \item \textbf{w/o contact ID in generation.}  All fingertips share a unified contact ID during grasp pose generation, meaning that the contact map feature of input point cloud does not contain any semantic information.
    \item \textbf{w/o contact ID in adaptation.}  All fingertips share a unified contact ID during both the classifier and adaptation models, meaning that the contact map feature of input point cloud does not contain any semantic information.
    \item \textbf{w/o adaptation.} The objects are directly lifted after the initial pose generation without adaptation.
    \item \textbf{Object-only.} Only object point cloud input with tactile mapping.
    \item \textbf{Full Point Cloud w/o tactile.} The combined hand-object point cloud is used, excluding all tactile-related features.
    \item \textbf{w/o PC.} Use images labeled with tactile signals instead of point clouds labeled with same tactile information. 
\end{itemize}

\input{table/ablation}

Table~\ref{ablation} shows the ablation results about our method. As shown in Table~\ref{ablation} (Row 1, Row 2, and Row 7), the semantic information provided by contact ID proves essential for both the initial hand pose generation and subsequent refinement. It enables the model to be aware of the current prediction or contact states, thereby producing more stable hand poses.
Moreover, the comparison between Row 3 and Row 7 highlights the critical role of the adaptation module, which effectively refines suboptimal initial poses. This improvement is especially notable on unseen objects (from 78\% to 82\%) and unseen categories (from 59\% to 83\%), demonstrating the strong robustness and generalization capability of our approach. Furthermore, the comparison between Row 4, Row 5 and Row 7 shows that without hand PC, we cannot have hand-object relation, and without tactile signals, the adaptation will turn harder. Additionally, the comparison between Row 6 and Row 7 demonstrates that without PC, the network cannot fully understand the geometric structure of objects through images. We further show hand pose generation performance empowered by contact id and adaptation performance in Figure~\ref{fig:generation} and Figure~\ref{fig:adaptation}. Figure~\ref{fig:generation} shows that under the guidance of the contact ID, the contact-driven grasp pose generator successfully produced initial poses that are geometrically consistent and meet the task requirements, and Figure~\ref{fig:adaptation} demonstrates the dynamic refinement process, where the adaptation module iteratively transforms initial unstable grasps into secure ones.


\begin{wrapfigure}{r}{0.495\textwidth} 
  \centering
  \includegraphics[width=0.5\textwidth]{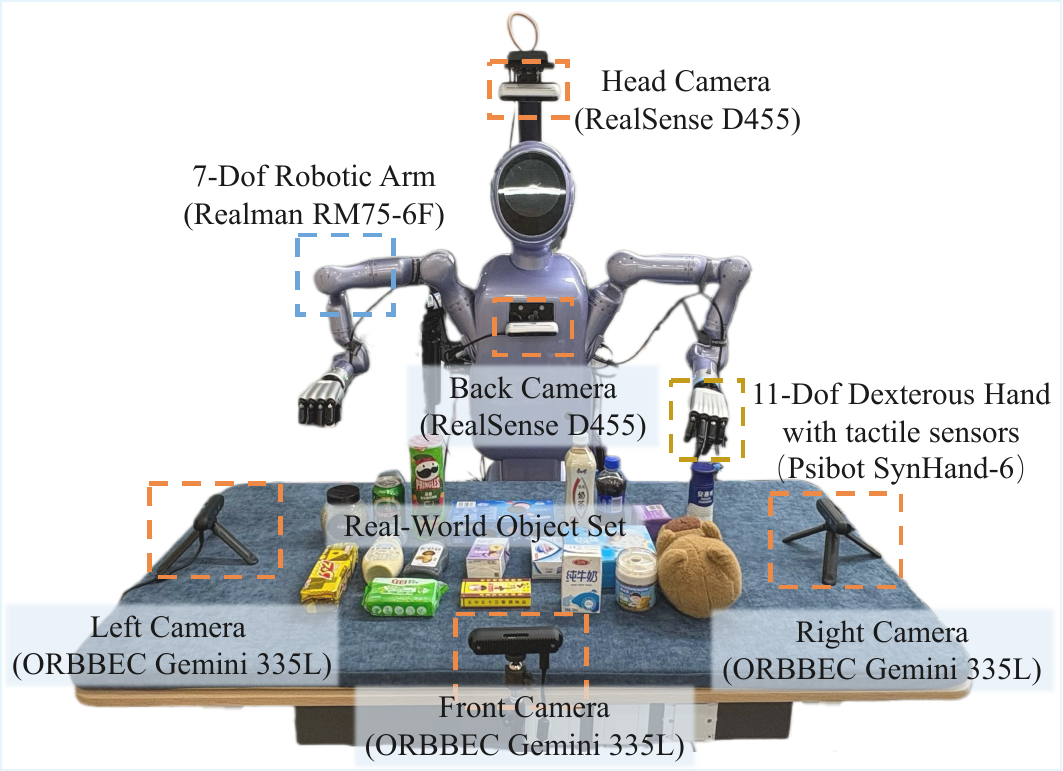}
  \caption{\textbf{Real-World Setup.}}
  \label{fig:real_world}
\end{wrapfigure}

\subsection{Real-World Experiment}
\label{real-world experiment}

\input{table/real-world}

To evaluate our approach in real-world settings, we conduct physical experiments using a Psibot robotic hand equipped with high-resolution tactile sensors. The real-world environment setting is shown in Figure~\ref{fig:real_world}.

We evaluate our method on more than 20 objects with distinct geometric and material properties. The evaluation metrics are consistent with those in simulation. A grasp is considered successful if the object remains stably held for at least three seconds after lifting. Table~\ref{real-world evaluation} reports the final evaluation results. Figure~\ref{fig:generation} and Figure~\ref{fig:adaptation} also illustrate the hand pose generation and adaptation performance in real world. More visualization results can be found in supplementary material.


\subsection{Representation Choice} 
Selecting an effective state representation is a critical trade-off between informational density and the feasibility of sim-to-real transfer. Our tactile-mapped point cloud, denoted as $\mathcal{P}_{vt} = \{(x_i, y_i, z_i, r_i, g_i, b_i, k_i, \mathbf{c}_i) \}$, simultaneously provides RGB, 3D geometry, and local contact semantics derived from tactile feedback.

While we explored richer tactile modalities such as force directions and shear forces, we found that the denser representation resulted in much greater data requirements, while our current tactile representation can already tackle most scenarios for dexterous grasping with less data. In small data tests, our accuracy and the accuracy when using dense modality were 88\% and 81\%, respectively. 
Moreover, choosing rich modalities places significant demands on both the real-world hardware and the sim-to-real gap of tactile representations, making it difficult to align the configurations in simulation and the real world. We also tested accuracy in real world, and the accuracy of our method and the method using the dense modality were 85\% and 72\%, respectively. Therefore, we eventually selected the relatively simple representations due to the effectiveness and efficiency, real-world and simulation alignment, achieving good experimental performance.


\subsection{Deep Analysis in Tactile Signals} 
\begin{figure*}[htb]
  \centering
  \includegraphics[width=1\linewidth]{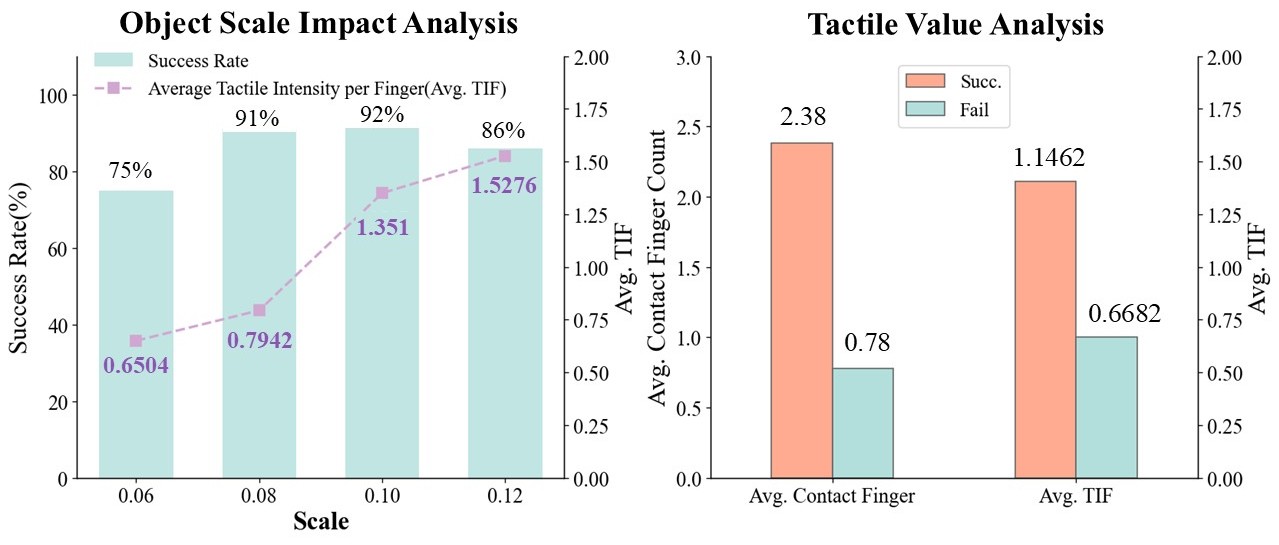}
  \caption{\textbf{(Left)} As the size of an object increases, Avg. TIF also increases, while grasping success rate first rises and then slightly decreases. This is because small objects are prone to slipping, whereas too large objects are difficult to grasp. \textbf{(Right)} The Avg. TIF and the number of contact fingers in successful grasping postures are higher than those in failed grasping postures.}
  \label{fig:onecol}
\end{figure*}

Beyond quantitative success rates, we analyze the physical interpretability of the learned contact semantics to verify their correlation with grasp stability. Specifically, since our tactile signals are normalized to a range of $[0, 1]$, we conducted detailed experiments to explore the relationship between tactile signals and stable grasping. Figure~\ref{fig:onecol} shows the detailed analysis.

%% file: table/baseline.tex
\begin{table}[ht]
  \centering
  \renewcommand{\arraystretch}{1.0}
  \setlength{\tabcolsep}{10pt}
  \caption{Comparison Results between Baselines and Our Method.}
  \label{baseline}
  \resizebox{0.95\textwidth}{!}{
  \begin{tabular}{cccc}
    \toprule
    Method & Seen Objects & Unseen Objects & Unseen Categories \\
    \midrule
    UniDexGrasp++ & 55\% & 49\% & 42\% \\
    DexGraspAnything & 77\% & 72\% & 67\% \\
    DexDiffuser & 71\% & 69\% & 55\% \\
    UniDexGrasp (PPO) & 52\% & 46\% & 40\% \\
    Robot Synesthesia & 33\% & 29\% & 22\% \\
    Intuitive Closed-Loop & 76\% & 71\% & 65\% \\
    ContactDexNet & 72\% & 68\% & 63\% \\
    DexGraspVLA & 69\% & 60\% & 58\% \\
    UniDexGrasp-CL & 59\% & 52\% & 46\% \\
    DexDiffuser-VT & 75\% & 71\% & 57\% \\
    Ours & \textbf{91\%} & \textbf{82\%} & \textbf{83\%} \\
    \bottomrule
  \end{tabular}
  }
  \vspace{-4mm}
\end{table}

%% file: table/ablation.tex
\begin{table}[htb]
  \centering
  \renewcommand{\arraystretch}{1.0}
  \setlength{\tabcolsep}{10pt}
  \vspace{-6mm}
  \caption{Ablation Results about Our Method.}
  \label{ablation}
  \resizebox{0.95\textwidth}{!}{
  \begin{tabular}{cccc} 
    \toprule
    Method & Seen Objects & Unseen Objects & Unseen Categories \\
    \midrule
    w/o contact id in adaptation & 84\% & 72\% & 73\% \\
    w/o contact id in generation & 79\% & 75\% & 69\% \\
    w/o adaptation & 81\% & 78\% & 59\% \\
    Object-only & 72\% & 67\% & 61\% \\
    Full Point Cloud w/o tactile & 78\% & 74\% & 67\% \\
    w/o PC & 74\% & 71\% & 64\% \\
    Ours & \textbf{91\%} & \textbf{82\%} & \textbf{83\%} \\
    \bottomrule
  \end{tabular}
  }
  \vspace{-6mm}
\end{table}

%% file: table/real-world.tex
\begin{table}[htbp]
  \centering
  \vspace{-6mm}
  \renewcommand{\arraystretch}{1.0}
  \setlength{\tabcolsep}{10pt}
  \caption{Real World Evaluation Results.}
  \vspace{-2mm}
  \resizebox{0.95\textwidth}{!}{
  \begin{tabular}{ccccc} 
    \toprule
    Method & Seen Objects & Unseen Objects & Unseen Categories \\
    \midrule
    DexGraspAnything & 81\% & 71\% & 73\% \\
    DexDiffuser & 77\% & 65\% & 52\% \\
    DexGraspVLA & 64\% & 57\% & 55\% \\
    Ours & \textbf{90\%} & \textbf{87\%} & \textbf{81\%} \\
    \bottomrule
  \end{tabular}
  }
  \label{real-world evaluation}
  \vspace{-4mm}
\end{table}

%% file: sec/5_Conclusions_and_Discussions.tex
\section{Conclusion}
In this work, we presented a unified visuo-tactile-fusion grasping framework that enables robust and generalizable dexterous grasping by tightly coupling tactile feedback with visual geometric perception. Through a contact-aware representation that binds tactile signals with finger identities and a tactile-guided adaptation mechanism for real-time refinement, our method enables fine-grained reasoning over finger–object interactions, enhancing both initial grasp feasibility and adaptive stability in simulation and real-world settings, further providing the research community with a feasible and scalable technical route for integrating visual and tactile modalities in dexterous manipulation.
